\documentclass[conference]{IEEEtran}
\IEEEoverridecommandlockouts

\usepackage{cite}
\usepackage{amsmath,amssymb}
\usepackage{graphicx}
\usepackage{booktabs}
\usepackage{rotating}
\usepackage{url}
\usepackage[hidelinks]{hyperref}
\usepackage{microtype}

\newcommand{\nTasks}{321}
\newcommand{\nPairs}{45}
\newcommand{\nPairTasks}{90}
\newcommand{\nSeeds}{5}
\newcommand{\nGrades}{14,445}
\newcommand{\nArchetypes}{9}
\newcommand{\corpusFingerprint}{35c002b563fc9c8f}
\newcommand{\pairReseed}{1.00}
\newcommand{\pairsVerified}{45}
\newcommand{\faultIntensity}{0.30}
\newcommand{\gateExact}{85.4}
\newcommand{\gatePassK}{85.4}
\newcommand{\gateMean}{31.37}
\newcommand{\gateUnresolvable}{245.16}
\newcommand{\optExact}{65.7}

\newcommand{\optPremature}{66.0}
\newcommand{\optSettleTwin}{113.07}
\newcommand{\optFailTwin}{393.99}
\newcommand{\optPairMean}{253.53}
\newcommand{\gatePairMean}{103.91}
\newcommand{\reactExact}{54.0}
\newcommand{\reactPassK}{17.4}
\newcommand{\ruleExact}{34.0}
\newcommand{\rulePassK}{6.9}
\newcommand{\rankInversions}{7}
\newcommand{\gateExactCI}{85.4 [79.9, 90.2]}
\newcommand{\gateGapCI}{31.37 [16.28, 48.27]}
\newcommand{\gateVsReact}{-96.81 [-118.46, -77.91]}
\newcommand{\bootReplicates}{2000}
\newcommand{\bootUnits}{276}
\newcommand{\tunedGap}{37.48}
\newcommand{\controlGap}{3.85}

\newcommand{\costConfigs}{45}
\newcommand{\costStable}{20}
\newcommand{\costGateAhead}{45}
\newcommand{\planSearchSize}{1,215}
\newcommand{\planSearchTasks}{40}
\newcommand{\planSearchBeaten}{2}
\newcommand{\oraclePremature}{86.0}
\newcommand{\lossFiring}{2.26}
\newcommand{\dupFiring}{1.49}
\newcommand{\reorderFiring}{0.68}
\newcommand{\dupMajority}{21.41}
\newcommand{\lossAlone}{77.28}
\newcommand{\delayAlone}{25.71}
\newcommand{\lossDelayBoth}{89.81}
\newcommand{\lossDelayInter}{13.18}

\newcommand{\nModelsBlocked}{2}
\newcommand{\blockedModels}{\texttt{gemma-4-26b-a4b-it}, \texttt{gemma-4-31b-it}}
\newcommand{\modelSubsetEpisodes}{86}
\newcommand{\gateOnSubsetExact}{93.0}
\newcommand{\gateOnSubsetGap}{6.30}

\newcommand{\bestModel}{gemini-3.6-flash}
\newcommand{\bestModelExact}{93.0}
\newcommand{\bestModelGap}{13.99}
\newcommand{\modelPremature}{0.4}
\newcommand{\modelProbeRate}{99.6}

\begin{document}

\title{FinalityBench: An Effect-Level Benchmark for Agent Decisions\\Under Delayed and Conflicting Financial Finality}

\author{\IEEEauthorblockN{Abhishek Sharma}
\IEEEauthorblockA{Senior Member, IEEE\\
\texttt{abhicse24@gmail.com} \quad ORCID: 0009-0007-1103-2103}}

\maketitle

\begin{abstract}
A merchant's payment processor, ledger, ERP and bank feed are updated by
messages that get delayed, duplicated, dropped and reordered, so for minutes at
a time the four hold contradictory beliefs about the same order. An agent
resolving the exception must decide whether to ship goods, re-submit a capture,
refund or wait, knowing some of those cannot be undone. We present
FinalityBench, an executable benchmark for that decision. It keeps a hidden
canonical event log and derives each system's view from a separately faulted
delivery stream, so disagreement follows from specified fault semantics rather
than being authored. Grading is on executed monetary effects: an episode is
scored by the merchant's terminal economic position, relative to a privileged
reference told when the pending capture resolves. The corpus of \nTasks{} tasks
includes \nPairs{} \emph{twin pairs} (\nPairTasks{} tasks): tasks whose four
system views are identical at the decision instant, whose authoritative probes
both return \texttt{unknown}, and whose eventual correct dispositions differ.
That snapshot indistinguishability is checked under every evaluation seed rather
than assumed; equivalence over all interaction traces is not claimed. Over \nGrades{} graded episodes from nine
programmatic policies, ranking by single-task accuracy and by paired loss
disagree in \rankInversions{} places: a ship-on-first-sign policy is second-best
by accuracy at \optExact\%\ and worst in the suite by paired loss, because it
cannot tell the two members apart. A runtime gating irreversible actions on
an authoritative finality probe reaches \gateExact\%\ and, unlike every polling
policy, loses nothing to pass\textsuperscript{\nSeeds}; its residual loss is
almost entirely one archetype, which prices finality information directly.
Language models reach the same exact rate as the hand-written gate on a stratified subset, lose about twice as much money, and discover the finality-gating strategy without being told it. \end{abstract}


\section{Introduction}

Ask an operations engineer what makes payment exceptions hard and the answer is
rarely that the rules are complicated. The rules are short. What makes them hard
is that the systems holding the facts disagree, and the disagreement is
temporary, and you cannot tell from inside a single system whether you are
looking at a settled truth or at a message that has not arrived yet.

Consider one order. The processor's replica says a capture is still submitted.
The ledger has posted the settlement twice. The ERP thinks the order is paid.
The bank statement shows the cash. Every one of those systems is behaving
correctly given what it was told. An agent asked whether to release the goods
has to decide which of them to believe, and shipping is not a decision it can
revisit.

Two properties of this setting are unusual enough to be worth a benchmark of
their own. First, some of the uncertainty is irreducible: a capture submitted
ninety seconds ago has no outcome yet, and no query resolves it, because there
is nothing to resolve. Waiting is the only thing that helps, and waiting costs a
delivery deadline. Second, the cost of being wrong is denominated in money and
is deeply asymmetric---shipping goods against a capture that later fails loses
the cost of those goods; re-submitting a capture that had in fact settled
charges the customer twice and buys a dispute.

Existing agent benchmarks measure neither. Benchmarks built on financial
documents grade a produced artifact against a static answer
key~\cite{finbalance}. Benchmarks built on executable enterprise stacks grade
functional correctness of a workflow~\cite{cfagentbench}. Benchmarks built
around delayed tools measure whether an agent uses the waiting time
well~\cite{asynctool}. None of them make \emph{finality}---whether an operation's
outcome has become a fact, as the independent variable, and none grade in dollars
of irreversible effect.

This paper contributes:

\begin{itemize}
\item \textbf{An environment} in which a hidden canonical event log is projected
into four operational systems through independently faulted delivery streams.
Six fault families are controlled by a profile, one is carried by the task
definition, and every projection is written the way that class of system is
actually built, so a given fault has a different observable signature depending
on where it lands (Section~\ref{sec:env}).

\item \textbf{A paired counterfactual construction} that makes premature
commitment measurably costly. Each pair is two tasks that no tool can
distinguish at the decision instant and whose correct actions differ; we verify
indistinguishability across every evaluation seed, and report what the construction required (Section~\ref{sec:pairs}).

\item \textbf{An effect-level monetary oracle} that scores the merchant's
terminal economic position from the canonical log, and reports the shortfall
against a reference that is told when the pending capture resolves and nothing
else. The reference is checked by exhaustive plan search
(Sections~\ref{sec:oracle} and~\ref{sec:results}).

\item \textbf{A reference evaluation} of nine programmatic policies over
\nGrades{} graded episodes, with fault-family attribution, a
tuning-transfer study against a regenerated split, and a sensitivity sweep over
the cost constants.
\end{itemize}

The model arm is deliberately narrow. It runs five models on a stratified subset that fits a free-tier quota, not the full corpus, and two of the five could not be evaluated at all because the provider's rate limit for them ended every episode before it finished. Section~
ef{sec:limits} says what that leaves unsettled.

\section{Related work}

\textbf{Financial agent benchmarks.} FinBalance~\cite{finbalance} reconciles
multi-document source bundles into journal entries and a balance sheet, scoring
against deterministic ground truth; its difficulty is accounting judgement over
static documents rather than asynchrony. CFAgentBench~\cite{cfagentbench} runs
agents against thirty-five self-hosted mock applications and grades functional
correctness, and notably includes a money-movement guard in which the correct
behaviour is to stage a payment for human approval. That guard is a static
policy rule: the action is forbidden regardless of system state. Ours is a
temporal condition---the same action is correct or catastrophic depending on
whether an outcome has become a fact.

\textbf{Delay, staleness and conflicting sources.} AsyncTool~\cite{asynctool}
introduces realistic tool latency and measures whether agents coordinate
concurrent work while waiting; the delay is a scheduling cost, not an epistemic
one, and there is no monetary consequence. ClawArena~\cite{clawarena} is the
closest in construction: each scenario keeps a complete hidden ground truth and
exposes only noisy, partial and sometimes contradictory traces. The difference
is what the agent produces. ClawArena grades multiple-choice selections and
shell checks---the agent reports a belief. FinalityBench grades money that
moved. An agent can hold an entirely correct belief here and still lose the cost
of goods by acting on it four hundred ticks too early. STALE~\cite{stale} asks
whether agents notice their memories have expired, which is the same question
one layer up from ours.

\textbf{Abstention and premature commitment.} Luo et al.~\cite{abstention} treat
abstention as a sequential decision in which an agent may answer, abstain, or
gather more information at each turn, and find that agents struggle when a task
looks feasible until the environment says otherwise. That is exactly the shape
of our unresolvable archetype, and our escalation action is priced rather than
free for the same reason their setting needs timing. Mehta~\cite{premature}
studies premature commitment as an \emph{interpretive} failure: an agent
settling on one reading of the evidence and defending it. Our premature action
is temporal: acting before the world has decided. The two can co-occur and are
not the same thing.

\textbf{Reversibility and verified tool calls.} Zhai et
al.~\cite{revisable} classify actions as idempotent, reversible, compensable or
irreversible and show that flexibility is bounded by reversibility; our action
space is built on that distinction. Mansoor et al.~\cite{verifiedtools} propose
a verification-aware tool wrapper for non-atomic tool failures, combining
postcondition checks, verify-before-retry and idempotency keys, and report that it reduces
duplicate actions. Our strongest baseline implements those three mechanisms and
adds a finality gate; Section~\ref{sec:results} separates what each contributes.
Cao et al.~\cite{corrupt} show that outcome-only scoring conceals procedural
violations in 27--78\% of reported successes, which is the general form of the
effect we measure in Section~\ref{sec:pairs}.

\textbf{Payment-specific formal work.} The lifecycle semantics our environment assumes, meaning when a capture is
determined and what a void or expiry does, are modelled formally in a companion artifact~\cite{holdspec}. Idempotent operation
identity for payment APIs, which is what our keyed-retry mechanism
depends on, is treated separately~\cite{t8}; work on generating retry-safe
payment services~\cite{t4} motivates the transactional baseline.

\section{The environment}
\label{sec:env}

\subsection{One truth, four beliefs}

A task is one order. Its hidden truth is an append-only log of events: order
placed, authorization approved, capture submitted, then a terminal outcome that
is either a settlement or a failure, plus an optional later chargeback. Nothing
in the benchmark ever shows an agent this log.

Each of the four systems receives a delivery stream derived from it. With no
faults enabled the streams differ only by a fixed per-system latency and the
bank's sixty-tick settlement batch, and the four systems agree as soon as those
pass. Enabling a fault family perturbs one specific property of the streams.

The folds differ, and the differences are the point:

\begin{itemize}
\item the \textbf{processor} is a status store keyed by operation, so writing
``settled'' twice is harmless and duplicate delivery is invisible there;
\item the \textbf{ledger} is an append-only journal that cannot recognise a
repeat, so a duplicate double-posts and a half-committed settlement leaves the
books out of balance;
\item the \textbf{ERP} is keyed by order and therefore idempotent;
\item the \textbf{bank} is a statement, so a duplicated settlement appears as two
lines and nothing in the feed says they are the same money.
\end{itemize}

A duplicate is therefore diagnosable only by comparing the ledger or bank
against the processor, and a partial commit only from inside the ledger. An
agent that reads one system cannot tell these apart, and four of our nine
archetypes punish trusting the processor while three punish trusting the ledger
or the bank.

\subsection{Fault families}

Six families are controlled by a fault profile and are the arms of the ablation:
delay, duplicate, loss, reorder, partial commit and stale read. Each fires with
probability equal to a single intensity parameter (\faultIntensity{} throughout)
so that no arm is confounded with a different tuning.

Two constraints on the engine were added after they caused problems.
Reordering exchanges the arrival times of two deliveries within a system, but
only when they are already within sixty ticks of each other and never when the
swap would place a delivery before the event it describes. Unbounded reordering, which the first version allowed, was inconsistent with
the delivery systems this environment models, and it silently destroyed the paired construction of
Section~\ref{sec:pairs} by revealing terminal outcomes early. A test caught the causality violation.

The seventh family, late reversal, is deliberately not profile-controlled. A
chargeback arriving after a correct decision is a new fact about the world, not
a mis-delivered old one, and it cannot be switched on for an existing canonical
log without changing the task. It is carried by the \texttt{late\_chargeback}
archetype and ablated by removing that archetype. An earlier version listed it
alongside the other six, where it fired exactly zero times in 1{,}605 episodes.

\subsection{Actions, and the limit of what any tool can do}

The agent has eleven tools. Four read a system's view. One,
\texttt{probe\_processor}, goes straight to the canonical log: it is immune to
every delivery fault and every read lag. The name is inherited from the
implementation and is misleading, because the call does not query the processor
replica at all; \texttt{query\_authoritative\_finality} would describe it
better. It is available to every policy at five ticks against a read's one, and
nothing prevents any policy from using it. Five tools are terminal
dispositions---ship, refund, write off, escalate, close. Two more advance the
clock or submit a capture.

The probe is the mechanism that makes the setting precise. Define an operation
as \emph{determined} at time $t$ when its terminal event has occurred at or
before $t$. The probe returns a status when the operation is determined and
\texttt{unknown} otherwise. That is not a limitation of the tool. It is the
statement that the outcome does not exist yet, and no sequence of calls
manufactures it. Only the clock does, and the clock is charged: every tool costs
simulated time, and shipping is impossible after the ship-by deadline because
the customer has cancelled.

\begin{table}[t]
\centering
\caption{The action space. Cost is in simulated ticks; the ship-by deadline is
480 ticks after the decision instant.}
\label{tab:tools}
\small
\setlength{\tabcolsep}{3.5pt}
\begin{tabular}{llr}
\toprule
tool & effect & cost \\
\midrule
\texttt{read\_processor} & processor replica & 1 \\
\texttt{read\_ledger}    & journal; appends every message & 1 \\
\texttt{read\_erp}       & order and fulfillment state & 1 \\
\texttt{read\_bank}      & settlement lines, batched & 1 \\
\texttt{probe\_processor} & authoritative; blank until determined & 5 \\
\texttt{wait}             & advance the clock & $n$ \\
\texttt{retry\_capture}   & submit under an idempotency key & 1 \\
\midrule
\texttt{ship}       & release goods, irreversible & 1 \\
\texttt{refund}     & return money, irreversible & 1 \\
\texttt{write\_off} & abandon collection & 1 \\
\texttt{escalate}   & hand to an analyst; not free & 1 \\
\texttt{close}      & close with no action & 1 \\
\bottomrule
\end{tabular}
\end{table}

\texttt{retry\_capture} takes an idempotency key. Reusing the original capture's
reference makes the call a no-op against a capture that already settled;
inventing a fresh key creates a second, genuinely new capture. The two calls are
indistinguishable at the moment they are made and differ by the order value
afterwards.

\subsection{A worked case}

Take a \texttt{lost\_settlement} task. The capture is submitted at $t=10$ and
settles at $t=300$; the settlement message to the processor and to the ERP is
dropped. The agent is handed the case at $t=120$ with a ship-by deadline at
$t=600$.

At $t=121$ the processor says the capture is submitted. The ERP says the order
is placed. The ledger, which was told about the submission, shows a clearing
balance. Nothing has settled anywhere. A policy that reads the processor and
stops concludes the money has not arrived; if it re-submits under a fresh key it
charges the customer a second time, because the original settles at $t=300$
regardless. Reading the bank feed instead gives a different answer, but not
until $t=360$, when the batch that covers $t=300$ posts.

The authoritative probe answers \texttt{unknown} at $t=121$ and \texttt{settled}
from $t=300$, because that is when the outcome starts existing. Waiting and
probing costs about a third of the deadline and gets the goods out on time. The
task is winnable, and four of our nine policies do not win it.

\subsection{Nominal parameters}

The numbers that set what counts as expensive are: cost of goods 60\% of order
value; dispute fee \$25; escalation \$18; step budget 40 calls; a read costs one
tick and an authoritative probe five. One tick is one second. A case is handed
over at $t=120$, must ship by $t=600$, and the world runs to $t=1800$. Order
values are drawn from eight retail amounts between \$24.99 and \$3,499.00, and
determination times are uniform on the interval each archetype specifies. These
are stated here rather than only in the artifact because they decide the
economic ranking; Section~\ref{sec:results} reports what survives moving them.

\subsection{Grading executed effects}
\label{sec:oracle}

After an episode ends the world runs on to the horizon, since events already in the log happen
whether or not the agent is still watching, and the merchant's
terminal position is computed as
\[
  \text{cash} - \text{cogs} - \text{liability} - \text{dispute} - \text{escalation},
\]
where cash is what actually settled in the canonical log, cogs is incurred only
if goods left, and liability is money held beyond the value of goods the
customer actually received. Liability covers both an over-collection on a
delivered order and a full collection on an order that never shipped, and it is
zero when the customer got what they paid for. A dispute fee applies whenever a
liability exists at all. The four deductions are disjoint, so a loss breakdown
sums to the total.

We report \emph{excess loss}: the reference position minus the achieved one.
The name is convenient but slightly wrong, and it is worth being precise about
why. The reference is privileged rather than optimal, so the quantity is a gap
to a strong comparator, not a distance below a proven upper bound, and it takes
negative values on tasks where a later fact punishes a decision that was correct
when it was made. Read it as an economic-position gap.
Some tasks contain losses no policy can avoid, and charging those to the agent
would rank policies by which tasks they drew. Subtracting a reference cancels
them exactly.

The reference is a privileged policy told one thing no channel provides, namely when and how the pending capture
resolves, and nothing else. It is not clairvoyant:
it does not know a chargeback is coming, so it ships an order a clairvoyant
policy would hold. That is the right bound for this benchmark, because the
question is whether an agent can act well under uncertain finality, not whether
it can foresee the future. It also means the gap can go negative on the rare tasks where a later fact
punishes a decision that was correct when it was made.

\section{Paired counterfactuals}
\label{sec:pairs}

The construction the rest of the paper leans on is this. A \emph{twin pair} is
two tasks such that, at the decision instant, all four system views are
identical up to renaming the case's own identifiers, both authoritative probes
return \texttt{unknown}, and the eventual correct dispositions differ: one
capture will settle, the other will fail unrecoverably.

What is verified is snapshot equality at the decision instant, not equivalence
over every interaction trace. A policy that waits, re-reads in a different
order, or issues a keyed retry could in principle observe a difference before
either capture resolves; we have not proved it cannot. The claim the experiments
rest on is the weaker one: from the four-system snapshot and an authoritative
probe at the decision instant, the members are indistinguishable. Establishing
trace equivalence would need an exhaustive search over legal action histories
that stop short of either determination time, and that is not done here.

Note also that "opposite correct actions" describes eventual dispositions. At
the decision instant the best first move may well be the same for both members:
wait.

A policy that commits at that instant cannot distinguish them. A policy that commits without
establishing finality therefore scores well on one and badly on the other, and
its pair mean prices the gamble instead of rewarding it. Aggregate accuracy does
the opposite: because settling outcomes are more common than failing ones, a
policy that always ships is rewarded for a bet that happens to pay off more
often than it does not.

Making this sound took four corrections, and
each looked like a different problem first.

\begin{enumerate}
\item The fault engine drew its random numbers only for subscribed
(event, system) pairs. A settlement is broadcast to four systems and a failure
to one, so the two members of a pair ran off different random streams. Drawing
before the subscription test fixed it.
\item Duplicate delivery offsets were drawn inside the emission loop, which runs
only for subscribed events, reintroducing the same desynchronisation one level
down and giving the two members different stale-read lags.
\item Reordering consumed the shared generator once per adjacent delivery pair,
and the two members have different numbers of deliveries. Keying the reorder
draw on the identities of the two deliveries rather than on their position made
it independent of how many others exist.
\item Comparing raw views always reported a difference, because the two tasks
necessarily carry different identifiers and those appear inside the views. The
property wanted is indistinguishability \emph{up to renaming of the case}.
\end{enumerate}

Before these fixes, requiring the property to hold across all \nSeeds{}
evaluation seeds left one valid pair out of forty-five, after an average of
forty-seven reseeding attempts. After them the mean number of attempts is
\pairReseed{}: the construction holds by design rather than by search, and all
\pairsVerified{} shipped pairs re-verify under an independent check.

\section{Corpus and policies}

\subsection{Corpus}

The corpus is \nTasks{} tasks over \nArchetypes{} archetypes, drawn from a fixed
mix and identified by a fingerprint (\texttt{\corpusFingerprint}) printed with
every result. Four archetypes punish believing the processor
(\texttt{lost\_settlement}, \texttt{stale\_processor}, \texttt{pending\_settle},
\texttt{unresolvable}); three punish believing the ledger or bank
(\texttt{duplicate\_settlement}, \texttt{partial\_ledger},
\texttt{late\_chargeback}). Chargebacks are rare on purpose---a rate anywhere
near the others would make refusing to ship the winning strategy, which is not
how card economics work.

A second split is \emph{regenerated} from a different master seed by the same
generator rather than held out from the same draw. Tuning against the public
split therefore cannot transfer through memorised task identities, only through
whatever the generator makes true of every draw.

\subsection{Policies}

Nine policies share one interface and are graded by identical code. In rough
order of sophistication: \emph{random}; \emph{optimistic ship}, which releases
goods as soon as the ledger shows a clearing entry, which is posted when a capture is
\emph{submitted} and long before anyone knows it will settle; \emph{eager}, which
reads the processor once and commits; \emph{majority vote}, which reads all four
systems and follows the majority; a \emph{rule-based SOP} with two tunable
polling knobs; a \emph{ReAct loop}, which is the observe--decide alternation
without a language model in it, included so that the contribution of the loop
shape can be separated from whatever reasons inside it; \emph{reflection}, which
pauses before anything irreversible and looks again; a \emph{finality gate},
which implements postcondition verification, keyed retries and
verify-before-retry~\cite{verifiedtools} and additionally refuses irreversible
action until an authoritative probe returns a terminal status; and the
\emph{finality oracle} reference.

Only the oracle is privileged, and the test suite asserts that.

\section{Experimental setup}

Every task is run under \nSeeds{} evaluation seeds; the seed is mixed with the
task's own schedule seed rather than replacing it. An earlier version replaced
it, and because every case's log has the same four-event shape, all \nTasks{}
tasks received an identical fault pattern---the same delivery lost in every
one. Three policies scored a flat zero and it looked like a policy bug for some
time.

The canonical log, delivery schedule and every grade are written to PostgreSQL
17 in Docker, with a JSONL fallback so a run never depends on a container. Each
experiment writes one JSON file carrying its numbers, the corpus fingerprint,
the seeds and the code revision; every table and figure in this paper is
generated from those files, and no number here was typed in by hand.

\section{Results}
\label{sec:results}

\subsection{The comparison}

Table~\ref{tab:main} gives the headline numbers over \nGrades{} graded episodes
and Table~\ref{tab:archetype} breaks accuracy down by archetype.

\begin{table*}[t]
\centering
\caption{Nine policies over \nTasks{} tasks and \nSeeds{} evaluation seeds
(\nGrades{} graded episodes). The gap is the shortfall against the
finality-oracle reference; ``pre-det.\ irrev.'' counts irreversible actions
taken before the pending capture's outcome existed. Brackets are 95\%
percentile bootstrap intervals over \bootUnits{} independent units, with twin
pairs resampled as one unit.}
\label{tab:main}
\small
\begin{tabular}{lrrrrrrr}
\toprule
& \multicolumn{2}{c}{exact (\%)} & \multicolumn{3}{c}{position gap (\$)} & \multicolumn{2}{c}{effects (\%)} \\
\cmidrule(lr){2-3}\cmidrule(lr){4-6}\cmidrule(lr){7-8}
policy & single run & pass\textsuperscript{5} & mean & p95 & max & pre-det.\ irrev. & duplicate \\
\midrule
random & 21.5 [16.9, 26.1] & 21.5 & 258.78 [212.00, 313.41] & 1,417.60 & 3,499.00 & 43.9 & 15.6 \\
eager & 14.0 [10.7, 17.4] & 14.0 & 298.84 [250.02, 344.99] & 1,424.60 & 1,424.60 & 0.0 & 0.0 \\
majority vote & 33.5 [29.3, 37.3] & 16.2 & 229.06 [191.66, 266.67] & 1,399.60 & 4,898.60 & 0.0 & 0.0 \\
optimistic ship & 65.7 [62.0, 69.2] & 32.7 & 163.61 [126.76, 205.37] & 999.00 & 3,499.00 & 66.0 & 0.0 \\
rule-based SOP & 34.0 [29.7, 38.1] & 6.9 & 204.54 [173.21, 235.45] & 802.60 & 1,442.60 & 0.0 & 0.0 \\
reflection & 53.5 [49.0, 57.9] & 17.4 & 129.61 [107.51, 154.87] & 777.60 & 1,442.60 & 0.0 & 0.0 \\
ReAct loop & 54.0 [49.4, 58.3] & 17.4 & 128.19 [106.44, 152.98] & 777.60 & 1,442.60 & 0.0 & 0.0 \\
\bfseries finality gate & 85.4 [79.9, 90.2] & 85.4 & 31.37 [16.28, 48.27] & 137.60 & 1,442.60 & 0.0 & 0.0 \\
finality oracle (reference) & 100.0 [100.0, 100.0] & 100.0 & 0.00 [0.00, 0.00] & 0.00 & 0.00 & 86.0 & 0.0 \\
\bottomrule
\end{tabular}
\end{table*}

\begin{table*}[t]
\centering
\caption{Exactly-correct rate (\%) by archetype. Columns, left to right:
duplicated settlement; a chargeback landing after a correct decision; a
settlement the processor never heard about; a half-posted journal entry; a
recoverable capture failure; an unrecoverable one; an ordinary pending capture;
a stale processor replica; and an outcome arriving after the ship-by deadline.
The finality gate is exact on six of the nine; its remaining loss is
concentrated in the last.}
\label{tab:archetype}
\small
\begin{tabular}{lrrrrrrrrr}
\toprule
policy & \rotatebox{55}{\small duplicate} & \rotatebox{55}{\small late chargeback} & \rotatebox{55}{\small lost settlement} & \rotatebox{55}{\small partial ledger} & \rotatebox{55}{\small pending fail} & \rotatebox{55}{\small fail, no retry} & \rotatebox{55}{\small pending settle} & \rotatebox{55}{\small stale replica} & \rotatebox{55}{\small unresolvable} \\
\midrule
random & 26 & 23 & 22 & 31 & 4 & 23 & 19 & 12 & 37 \\
eager & 0 & 0 & 0 & 0 & 0 & 100 & 0 & 0 & 50 \\
majority vote & 27 & 35 & 0 & 46 & 0 & 100 & 44 & 28 & 50 \\
optimistic ship & 85 & 88 & 89 & 88 & 0 & 35 & 86 & 81 & 50 \\
rule-based SOP & 59 & 58 & 0 & 52 & 19 & 61 & 61 & 10 & 0 \\
reflection & 79 & 75 & 56 & 66 & 12 & 65 & 84 & 56 & 0 \\
ReAct loop & 79 & 75 & 56 & 66 & 16 & 65 & 84 & 56 & 0 \\
finality gate & 100 & 100 & 100 & 100 & 84 & 92 & 100 & 100 & 0 \\
finality oracle (reference) & 100 & 100 & 100 & 100 & 100 & 100 & 100 & 100 & 100 \\
\bottomrule
\end{tabular}
\end{table*}

The finality gate reaches \gateExact\%\ exact with a mean excess loss of
\$\gateMean. Its per-archetype breakdown is more informative than its mean: it
is exact on six archetypes and loses \$\gateUnresolvable{} per case on
\texttt{unresolvable}, the tasks whose outcome lands after the ship-by deadline.
That number is the price of not being able to see finality. No unprivileged
policy can guarantee avoiding it, though several avoid it on individual tasks by
gambling and winning, which is why Table~\ref{tab:archetype} shows non-zero
exactness there for policies that ship early.

\subsection{How much of this is sampling noise}

The \nGrades{} graded episodes are not \nGrades{} independent observations. The
same \nTasks{} tasks recur under every seed, a policy's behaviour on one seed is
correlated with its behaviour on another, and the two members of a twin pair are
two views of one construction rather than two draws. Intervals are therefore
computed by a bootstrap over tasks, resampling \bootUnits{} independent units
with twin pairs held together, \bootReplicates{} replicates.

The intervals are wide. The finality gate's mean gap is \$\gateGapCI, a range
of roughly threefold, and its exact rate is \gateExactCI\%. Point estimates on
a corpus this size carry that much uncertainty, and comparisons are the more
useful quantity: measured on a shared task resample, the gate's mean gap sits
\$\gateVsReact{} below the ReAct loop's, an interval that excludes zero, as does
every comparison between the gate and an ungated policy. The ReAct and
reflection loops are not separable at this sample size, which is consistent with
what Section~\ref{sec:results} says about reflection over a stale replica.

\subsection{Accuracy is not reliability}

\begin{figure}[t]
\centering
\includegraphics[width=\columnwidth]{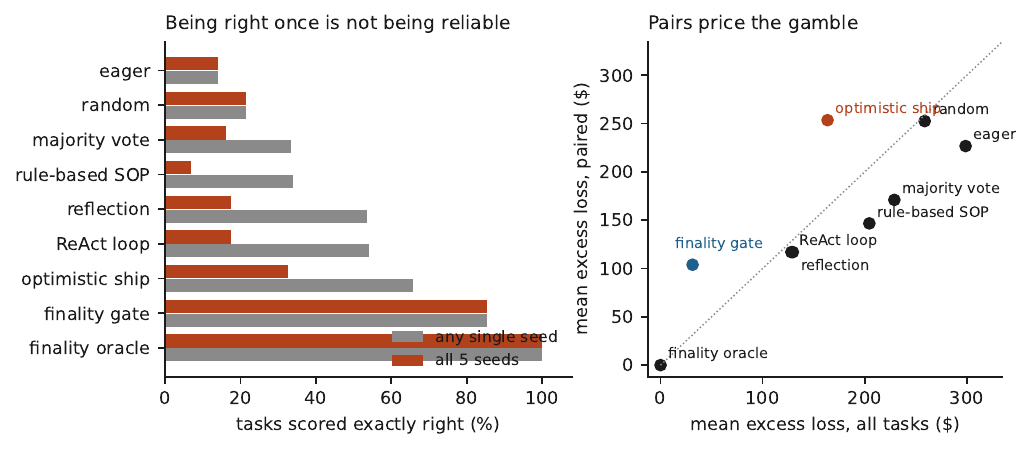}
\caption{Left: being right once against being right every time. Right: mean
excess loss over all tasks against mean over twin pairs; points above the
diagonal are policies that aggregate scoring flatters.}
\label{fig:policies}
\end{figure}

Figure~\ref{fig:policies} puts single-seed accuracy against
pass\textsuperscript{\nSeeds}, the share of tasks a policy gets right under
\emph{all} \nSeeds{} seeds. The polling policies collapse: the ReAct loop falls
from \reactExact\%\ to \reactPassK\%, the rule-based procedure from
\ruleExact\%\ to \rulePassK\%. The finality gate does not move at all
(\gateExact\%\ to \gatePassK\%).

The reason is structural rather than statistical. A policy that decides from
replica reads is correct when the message it needed happened to arrive, so its
correctness is a property of the fault draw. A policy that gates on the
authoritative probe either can establish finality before the deadline or cannot,
and which delivery was dropped does not enter into it. Reporting a mean over
seeds would have hidden this entirely.

\subsection{What the pairs reveal}

\begin{figure}[t]
\centering
\includegraphics[width=\columnwidth]{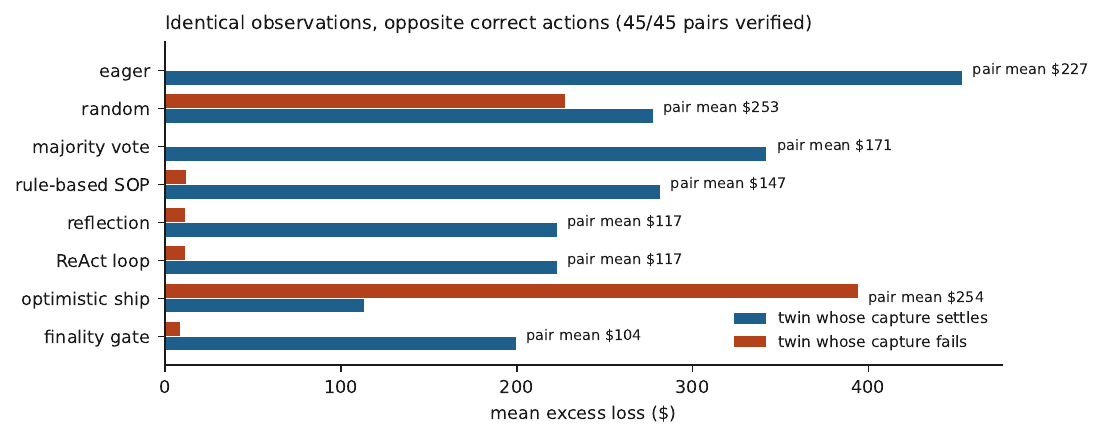}
\caption{The two halves of each pair. Aggregate scoring adds them together;
paired scoring averages inside the pair, which prices the gamble.}
\label{fig:paired}
\end{figure}

Table~\ref{tab:paired} and Figure~\ref{fig:paired} split each pair into its two
halves.

\begin{table}[t]
\centering
\caption{Twin pairs: identical observations, opposite correct actions.
Columns are mean excess loss on the twin whose capture settles, on the twin
whose capture fails, the mean of the two, and the share of pairs on which an
irreversible action was taken before the outcome existed.}
\label{tab:paired}
\small
\setlength{\tabcolsep}{3pt}
\begin{tabular}{lrrrr}
\toprule
policy & settles (\$) & fails (\$) & pair (\$) & prem.\ (\%) \\
\midrule
random & 277.53 & 227.65 & 252.59 & 75.6 \\
eager & 453.46 & 0.00 & 226.73 & 0.0 \\
majority vote & 342.04 & 0.00 & 171.02 & 0.0 \\
optimistic ship & 113.07 & 393.99 & 253.53 & 63.1 \\
rule-based SOP & 281.70 & 11.68 & 146.69 & 0.0 \\
reflection & 222.66 & 11.28 & 116.97 & 0.0 \\
ReAct loop & 222.66 & 11.28 & 116.97 & 0.0 \\
finality gate & 199.42 & 8.40 & 103.91 & 0.0 \\
finality oracle (reference) & 0.00 & 0.00 & 0.00 & 100.0 \\
\bottomrule
\end{tabular}
\end{table}

The optimistic policy is the case worth stating plainly. By single-task accuracy
it is second-best in the suite at \optExact\%, behind only the finality gate. By
paired loss it is the worst policy we evaluated, at \$\optPairMean{} against the
gate's \$\gatePairMean. It collects \$\optSettleTwin{} on the twins whose
captures settle and \$\optFailTwin{} on the twins whose captures fail, and it
cannot tell which is which. It takes an irreversible action before the outcome
exists on \optPremature\%\ of tasks. Ranking the nine policies by accuracy and by
paired loss disagrees in \rankInversions{} places.

This is the concrete form of the general point that outcome-only scoring
conceals procedural failure~\cite{corrupt}. Here the concealment is not
incidental: because settling outcomes outnumber failing ones in any realistic
mix, aggregate accuracy \emph{systematically} rewards the bet.

\subsection{Language models}

\begin{table}[t]
\centering
\caption{Models on a stratified subset of \modelSubsetEpisodes{} episodes, with
the finality gate scored on exactly those episodes. ``Failed'' counts episodes
that ended without a decision, from a protocol error or a provider rate limit.}
\label{tab:models}
\footnotesize\setlength{\tabcolsep}{3pt}
\IfFileExists{tbl_models.tex}{\begin{tabular}{llrrrrrr}
\toprule
system & tier & episodes & failed & exact (\%) & mean gap (\$) & pre-det.\ (\%) & steps \\
\midrule
finality gate & policy & 86 & 0 & 93.0 & 6.30 & 0.0 & --- \\
\midrule
\texttt{gemma-4-26b-a4b-it} & open & 0 & 86 & \multicolumn{4}{c}{\emph{rate-limited, not evaluated}} \\
\texttt{gemma-4-31b-it} & open & 1 & 85 & \multicolumn{4}{c}{\emph{rate-limited, not evaluated}} \\
\texttt{gemini-3.1-flash-lite} & hosted-small & 86 & 0 & 64.0 & 82.83 & 0.0 & 11.6 \\
\texttt{gemini-3.6-flash} & hosted-mid & 86 & 0 & 93.0 & 13.99 & 0.0 & 15.1 \\
\texttt{gemini-3.1-pro-preview} & frontier & 86 & 0 & 87.2 & 13.82 & 1.2 & 13.3 \\
\bottomrule
\end{tabular}}{\emph{model results absent: run \texttt{make models}}}
\end{table}

The model arm runs on a stratified subset rather than the full corpus. Every
archetype is represented and complete twin pairs are kept, so the paired metric
stays computable, and the corpus is small enough to fit a free-tier quota. The
gate's row in Table~\ref{tab:models} is recomputed on exactly the episodes the
models attempted, because its full-corpus figure comes from a different and
harder task set; comparing the two would compare task sets rather than systems.

\bestModel{} reaches \bestModelExact\%\ exact against the gate's
\gateOnSubsetExact\%\ on the same episodes, at a mean gap of \$\bestModelGap{}
against \$\gateOnSubsetGap. The models match the hand-written policy on how
often they are exactly right and lose roughly twice as much when they are not.

Three observations, none of which we anticipated.

\textbf{The models find the authoritative channel unprompted.} Every evaluated
model used \texttt{probe\_processor} in \modelProbeRate\%\ of episodes and
waited in nearly all of them. The system prompt says a submitted capture has no
outcome until the processor produces one; it does not say which tool to prefer,
or that waiting is the remedy. The finality-gating strategy is discoverable
from the environment.

\textbf{They err toward caution rather than commitment.} Pre-determination
irreversible actions occur in \modelPremature\%\ of model episodes. On twin
pairs the asymmetry runs opposite to the optimistic policy's: models give up
margin on the twin whose capture settles rather than shipping goods against the
twin whose capture fails. On this corpus their failure mode is not the
premature commitment the benchmark was built to catch.

\textbf{The failures concentrate where the gate's do.} \bestModel{} records a
zero gap on seven of the nine archetypes, including the lost settlement, the
duplicated posting, the half-committed ledger and the stale replica. What it
loses, it loses on \texttt{pending\_fail}, where recovery needs a keyed retry,
and on \texttt{unresolvable}. That is close to the gate's own profile, which
suggests the two are limited by the same thing.

\subsection{Which faults cost what}

\begin{figure}[t]
\centering
\includegraphics[width=\columnwidth]{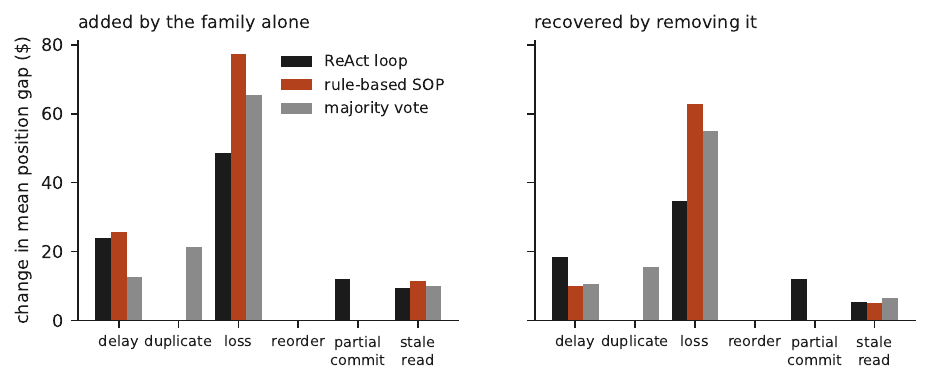}
\caption{Each fault family measured twice: what it costs on its own above a
fault-free baseline, and what removing it from the full profile recovers.
Duplicate delivery moves only majority vote, which reads the ledger and the
bank feed; reordering moves nothing.}
\label{fig:ablation}
\end{figure}

Table~\ref{tab:ablation} and Figure~\ref{fig:ablation} isolate each family and
remove each family, and report how often each actually fires.

\begin{table*}[t]
\centering
\caption{Fault attribution, in dollars of mean excess loss. ``Only'' is the
family alone above a fault-free baseline; ``minus'' is what removing it recovers
from the full profile. Firing rates are deliveries affected per episode.}
\label{tab:ablation}
\small
\begin{tabular}{lrrrrrrrr}
\toprule
& \multicolumn{2}{c}{majority vote} & \multicolumn{2}{c}{rule-based SOP} & \multicolumn{2}{c}{ReAct loop} & \multicolumn{2}{c}{finality gate} \\
\cmidrule(lr){2-3}\cmidrule(lr){4-5}\cmidrule(lr){6-7}\cmidrule(lr){8-9}
fault family & only & minus & only & minus & only & minus & only & minus \\
\midrule
none (baseline) & 129.89 & --- & 109.74 & --- & 45.52 & --- & 31.37 & --- \\
delay (1.52/ep) & 12.73 & 10.62 & 25.71 & 10.03 & 23.91 & 18.43 & 0.00 & 0.00 \\
duplicate (1.49/ep) & 21.41 & 15.46 & 0.00 & 0.00 & 0.00 & 0.00 & 0.00 & 0.00 \\
loss (2.26/ep) & 65.52 & 54.85 & 77.28 & 62.79 & 48.51 & 34.74 & 0.00 & 0.00 \\
reorder (0.68/ep) & 0.00 & 0.00 & 0.00 & 0.00 & 0.00 & 0.00 & 0.00 & 0.00 \\
partial commit (0.17/ep) & 0.00 & 0.00 & 0.00 & 0.00 & 12.05 & 11.88 & 0.00 & 0.00 \\
stale read (1.22/ep) & 10.11 & 6.48 & 11.38 & 4.99 & 9.52 & 5.31 & 0.00 & 0.00 \\
\midrule
all six & 229.06 & --- & 204.54 & --- & 128.19 & --- & 31.37 & --- \\
\bottomrule
\end{tabular}
\end{table*}

Message loss dominates, firing \lossFiring{} times per episode and accounting
for most of the damage on its own.

The two families that fire often and change little are worth separating
carefully, because an earlier version of this paper got the statement wrong.
Reordering fires \reorderFiring{} times per episode and changes no evaluated
policy's decision. Duplicate delivery fires \dupFiring{} times and changes
decisions for exactly one: majority vote, which reads the ledger and the bank
feed and is therefore exposed to double-posting, pays \$\dupMajority{} for it.
Policies that decide from processor status are unaffected, because a status
store keyed by operation cannot tell a repeat from a first delivery and does not
need to. So duplication is not inert; it is inert with respect to a particular
way of reading the systems.

Loss and delay interact, and establishing that needs the two-by-two rather than
a comparison of ``only'' against ``minus''. The latter shows non-additivity
between one family and the whole remaining profile, which is a weaker and
different statement. With both cells measured, the interaction is negative:
for the rule-based procedure, loss alone costs \$\lossAlone, delay alone
\$\delayAlone, and the two together \$\lossDelayBoth---\$\lossDelayInter{}
less than their sum. Once a message has been dropped the policy is already
without the information it needed, and delaying a second one adds less harm
than it would have on its own. The effect is in the same direction for the
ReAct and reflection loops.

The finality gate's mean position gap is identical in every arm. This is not a
discovery: a policy that decides only from the authoritative channel is
invariant to delivery faults by construction, because delivery faults are
applied to the channels it does not read. The ablation confirms the
implementation rather than establishing a result. What the arm does show is how
much the replica-reading policies lose to each family, and it is against those
numbers that the gate's flat line should be read.

That also bounds what this benchmark currently demonstrates. It shows the value
of an authoritative finality signal; it does not yet show an agent reasoning
well across conflicting evidence, because the winning policy avoids the
conflicting evidence entirely. Arms with no probe, a delayed or rate-limited
probe, or a probe that confirms operation identity without confirming
settlement would separate those two things, and they are not in this version.

\subsection{Does tuning survive a regenerated split?}

We grid-searched the rule-based procedure's two knobs on the public split and
re-measured on the regenerated one (Table~\ref{tab:transfer}).

\begin{table}[t]
\centering
\caption{Tuning transfer, in mean excess loss. ``Hidden'' is the regenerated
split. The finality gate, which has nothing fitted to either split, gives the
split-to-split floor.}
\label{tab:transfer}
\small
\setlength{\tabcolsep}{4pt}
\begin{tabular}{llrrr}
\toprule
configuration & knobs & public (\$) & hidden (\$) & gap (\$) \\
\midrule
rule-based, tuned & $(45,10)$ & 191.83 & 229.30 & 37.48 \\
rule-based, default & $(60,8)$ & 204.54 & 245.81 & 41.26 \\
finality gate (control) & --- & 31.37 & 35.22 & 3.85 \\
\bottomrule
\end{tabular}
\end{table}

Tuning improved the public split and the improvement carried over: the tuned
configuration is better than the default on the regenerated split too. What did
not carry is the absolute level. The tuned policy degrades by \$\tunedGap{}
across splits while the unfitted control degrades by \$\controlGap. the fitted knobs encode a property of the generator, poll more often, rather
than an identity memorised from the public draw. The control's own gap of
\$\controlGap{} is a single observation from one policy on one regenerated
split, so it indicates the scale of split-to-split variation without estimating
it; several regenerated corpora would be needed for that.

\subsection{Do the cost constants decide the answer?}

Three numbers set what the oracle calls expensive: cost of goods, dispute fee,
and the cost of handing a case to a human. They are defensible rather than
measured, so we swept them over \costConfigs{} configurations. The full
nine-policy ordering is identical in \costStable{} of them. The middle of the ranking is not stable,
and we do not claim it is. The finality gate is ahead of
every ungated policy in \costGateAhead{} of \costConfigs. The headline
conclusion is robust to these constants; fine-grained ordering among the middling
policies is not.

\subsection{Is the reference actually optimal?}

Excess loss is meaningless if some plan beats the reference, so we searched
exhaustively rather than asserting it: \planSearchSize{} plans of the form
\emph{wait, optionally retry under one of two keys, wait, take one terminal
action} on each of \planSearchTasks{} sampled tasks.

The first run found the reference beaten on eight tasks. Six were genuine
defects in it. On unresolvable cases whose capture will settle, the reference
was refunding when it should have shipped before the deadline, since the goods can go out now
and the money lands later. On a case whose capture will fail
recoverably, it waited for the failure it already knew was coming instead of
submitting the replacement immediately. Both were fixed. After the fix the
reference is beaten on \planSearchBeaten{} of \planSearchTasks{} tasks, both
\texttt{late\_chargeback}, the documented case in which a correct decision is punished by a fact that had
not happened when it was made.

One diagnostic is worth reporting for its own sake. The reference takes an
irreversible action before the outcome is determined on \oraclePremature\%\ of
tasks. It is entitled to, since that is what being told the outcome means, and the
figure measures how much of the reference's advantage comes from information no
observer has.

\section{Discussion}

Three things follow from these numbers.

\textbf{Authoritative versus replica-based observation.} Among the hand-written policies evaluated here, the gap between the
ReAct loop and the finality gate is not a difference in deliberation; both are
short procedures with fixed rules. It is that one asks the authoritative
channel and the other reads replicas. The experiment varies the channel far
more than it varies reasoning ability, so this is a statement about these
policies and not about reasoning in general. Reflection illustrates the same point from the other
side: re-reading the same replica in the same instant returns the same answer,
so a reflection step only helps if the clock moves first, and even then it
repairs mistakes caused by looking too early rather than by looking in the wrong
place. Its numbers are within noise of plain ReAct.

\textbf{Effect of retry-key selection on duplicate charges.} The keyed retry makes a
duplicate charge something a policy chooses rather than suffers. Policies that
reuse the original reference never produce a duplicate effect in our runs;
policies that mint a fresh key do so whenever the original settles after all.
This is the operational content of treating operation identity
formally~\cite{t8}.

\textbf{Residual loss when finality arrives after the deadline.} The
finality gate's residual \$\gateUnresolvable{} per unresolvable case is not a
defect of the policy. It is what the deadline costs when the outcome arrives
after it, and it is the same for every unprivileged policy. Separating that
component from avoidable loss is one of the things an effect-level oracle buys.

\section{Limitations}

\textbf{The model arm is narrow.} Five models were attempted on a
\modelSubsetEpisodes-episode stratified subset. \nModelsBlocked{} of them,
\blockedModels, could not be evaluated: the provider's free-tier rate limit for
those models ended almost every episode before it reached a decision, so the
open-weight tier is missing entirely and nothing here speaks to it. One vendor,
one prompt, one temperature setting, and two seeds is not a study of model
capability; it establishes that the benchmark runs against models, discriminates
between them, and that the strategy it rewards is discoverable.

\textbf{The earlier version of this paper evaluated no model at all.} The adapter in
\texttt{llm\_agent.py} implements the interface---observation rendering, tool schema and a strict reply parser. It is exercised
end to end against recorded transcripts in the test suite, but it has never called a model, because no model
endpoint was available in the environment where these results were produced. It
raises rather than falling back to a heuristic. Everything reported here is a
property of the benchmark and of hand-written policies; nothing here is evidence
about any language model.

\textbf{The policies are ours.} A hand-written ReAct-shaped procedure is an
ablation of the control structure, not a stand-in for an agent that reasons. It
bounds what the loop shape alone achieves and nothing more.

\textbf{The world is simulated.} No real processor was contacted. The lifecycle
the environment assumes is drawn from a formal model of provider
behaviour~\cite{holdspec}, but the delivery semantics, latencies and batch
cadence are chosen to be plausible rather than measured against a production
system, and real processors will differ.

\textbf{The economics are stipulated.} Cost of goods, the dispute fee and the
escalation cost are defensible round numbers, not estimates of any merchant's
books. Section~\ref{sec:results} reports what survives moving them, and the
middle of the policy ranking does not.

\textbf{The corpus is one shape.} Every task is a single order with one pending
capture. Multi-item orders, partial captures, split shipments and concurrent
exceptions on one customer are all absent, and each would add failure modes this
corpus cannot express.

\section{Conclusion}

FinalityBench makes the temporary disagreement between a merchant's payment
systems into an experimental variable, and grades agent decisions by the money
that actually moved. Its central construction is a pair of tasks that no tool
can tell apart and that demand opposite actions; scoring on those pairs turns
out to reorder the policy ranking in \rankInversions{} places relative to
aggregate accuracy, and to expose a policy that looks second-best as the worst
in the suite. A runtime that gates irreversible actions on an authoritative
finality probe is both the most accurate and the only one whose accuracy
survives repetition, and the loss it cannot avoid puts a price on finality
information itself.

The benchmark, the generator, the fault engine and the model adapter are
released so that the arm this paper leaves unrun can be run.

\section*{Artifact availability}

Code, corpus generator, Docker setup and every result file are at
\url{https://github.com/abhisheksharma2411/finalitybench}, archived at
\textsc{doi} \texttt{10.5281/zenodo.22262591}. \texttt{make reproduce} runs the
suite end to end from a clean tree. The corpus used here has fingerprint
\texttt{\corpusFingerprint}.

\end{document}